%% file: main.tex
\documentclass[10pt,twocolumn,letterpaper]{article}

\usepackage[pagenumbers]{cvpr} 
\usepackage{multirow}
\usepackage{graphicx}


\definecolor{cvprblue}{rgb}{0.21,0.49,0.74}
\usepackage[pagebackref,breaklinks,colorlinks,allcolors=cvprblue]{hyperref}

\def\paperID{*****} 
\def\confName{CVPR}
\def\confYear{2025}

\title{Towards Zero-Shot \& Explainable Video Description\\by Reasoning over Graphs of Events in Space and Time}

\author{Mihai Masala and Marius Leordeanu\\
Institute of Mathematics of Romanian Academy\\
{\tt\small mihaimasala@gmail.com, leordeanu@gmail.com}
}

\begin{document}
\maketitle
\input{sec/0_abstract}    
\input{sec/1_intro}

\input{sec/2_related}
\input{sec/3_method}

\input{sec/4_experiments}

\input{sec/5_results}
\input{sec/6_conclusions}

{
    \small
    \bibliographystyle{ieeenat_fullname}
    \bibliography{main}
}


\end{document}

%% file: sec/0_abstract.tex
\begin{abstract}
In the current era of Machine Learning, Transformers have become the de facto approach across a variety of domains, such as computer vision and natural language processing. Transformer-based solutions are the backbone of current state-of-the-art methods for language generation, image and video classification, segmentation, action and object recognition, among many others. Interestingly enough, while these state-of-the-art methods produce impressive results in their respective domains, the problem of understanding the relationship between vision and language is still beyond our reach. In this work, we propose a common ground between vision and language based on events in space and time in an explainable and programmatic way, to connect learning-based vision and language state of the art models and provide a solution to the long standing problem of describing videos in natural language. We validate that our algorithmic approach is able to generate coherent, rich and relevant textual descriptions on videos collected from a variety of datasets, using both standard metrics (e.g. Bleu, ROUGE) and the modern LLM-as-a-Jury approach. 
\end{abstract}

%% file: sec/1_intro.tex
\section{Introduction}
\label{sec:intro}

The task of describing the visual content of a given video in natural language, video captioning~\cite{wang2018reconstruction,liu2018sibnet,aafaq2019spatio,li2021value,lin2022swinbert,chen2023valor}, represents a challenge for both the computer vision and natural language processing communities.

Although there is a plethora of methods both from the field of video understanding (object detection and tracking~\cite{wang2023yolov7}, semantic segmentation~\cite{zou2023segment,cheng2021mask2former} and action recognition~\cite{tong2022videomae, wang2023videomae}) and that of natural
processing (LLM such as ChatGPT~\cite{hurst2024gpt}), we are still far from understanding how to best bridge the two fields and are still not able to describe in rich natural language the content of videos.

Before the recent rise of Visual Large Language Models (VLLMs), existing deep learning methods trained for video description are only able to produce very short captions of videos, being rather close to video classification (where a video could belong to a finite number of classes) than to that of describing in natural language such videos, with rich textual descriptions that could have infinitely many forms. Moreover, such models suffer from overfitting such that once given a video from an unseen context or distribution the quality and accuracy of the description drops, as our evaluations prove. On the other hand, VLLMs have shown impressive results, being capable of generating long, rich descriptions of videos. Unfortunately VLLMs still share some of the same weaknesses as previous methods: they are largely unexplainable and they still rely on sampling frames to process a video. Moreover, top-performing models such as GPT, Claude or Gemini are not open and  are only accessible via an paid API.

We argue that one of the main reasons why this interdisciplinary cross-domain task is still far from being solved is that we still lack an explainable way to bridge this apparently insurmountable gap. Explainability could provide a more analytical and stage-wise way to make the transition from vision to language that is both trustworthy and makes sense. It is clear that language is grounded in vision, as it describes events happening in the real world and being connected spatially, temporally and semantically, in which objects perform actions and interact in physical or semantic context that could be captured by vision and described by language.

\begin{figure*}
\includegraphics[width=1.0\linewidth]{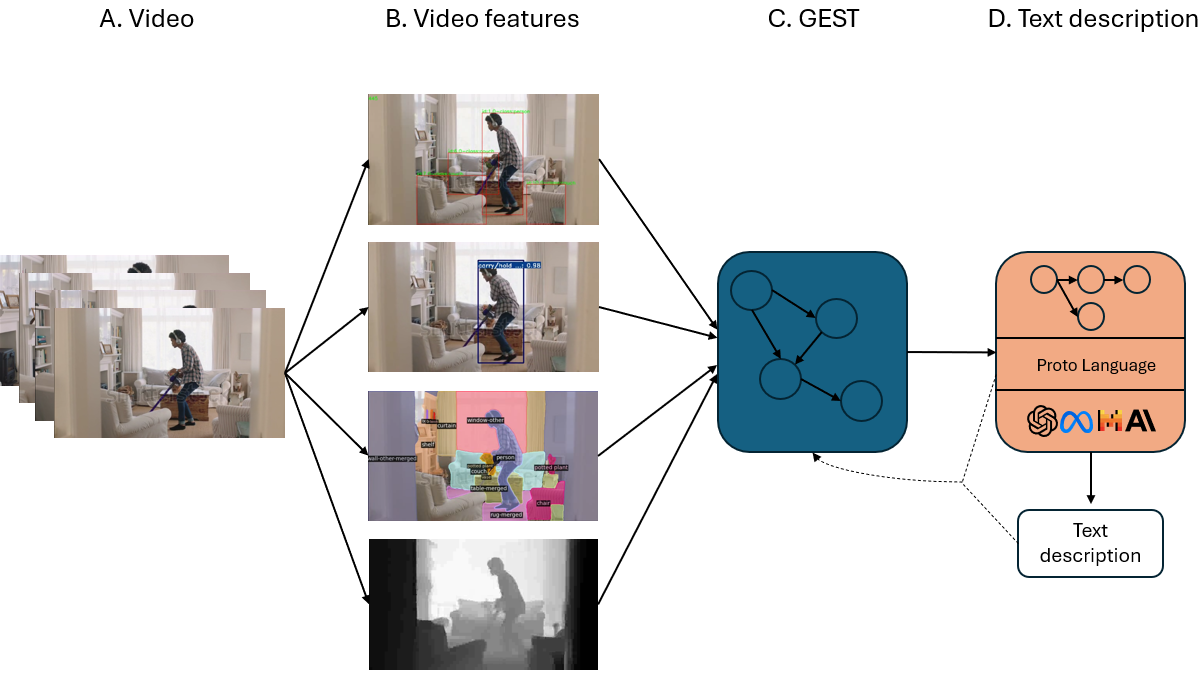}
\caption{An overview of our approach. Starting from a raw video we perform object detection and tracking, action detection, semantic segmentation and depth estimation. We aggregate this information to build the corresponding Graph of Events in Space and Time. By reasoning over (e.g., temporally and spatially sorting the graph, describing the events) this graph we build an intermediate representation in the form of a proto language. We prompt existing LLMs to take this proto language and transform it in a fully fledged natural, rich and accurate textual description. Furthermore, trusting LLMs with enough power to alter certain parts of the events (e.g., a miss-identified object) and learning from this process allows us to update the graph in order to obtain a more context-aware and accurate representation.
}
\label{fig:architecture}
\end{figure*}

In some sense, language "speaks" about what vision "sees" and it makes sense to think that vision comes first and then is followed by language, an observation that is in agreement with neuroscience studies about human brain development in infants. Given that today's learning models in both vision and language are so impressive, we believe that it is time to fully exploit such existing methods and create procedural methods that can offer the explainable bridge currently so much needed between vision and language. While learning novel vision-language models from data is both important and powerful, the direct path from vision to language by building procedures from the existing state of the art in both fields is left unexplored. Our proposed approach harnesses existing strong pre-trained vision models for a variety of tasks (i.e., action detection, object detection and tracking, semantic segmentation and depth estimation) to build an explicit, grounded representation in the form of a Graph of Events in Space and Time - GEST~\cite{masala2023explaining}. Furthermore, this representation is used to build an intermediate textual description (proto-language) that is then converted into a fully fledged rich textual description using text-only LLMs. An overview of our proposed approach is presented in Figure~\ref{fig:architecture}.

%% file: sec/2_related.tex
\section{Related Work}


Up until recently, most video captioning models were based on the encoder decoder architecture, using mostly CNNs for encoding the video frames and LSTMs to generate the textual description~\cite{wang2018reconstruction,liu2018sibnet}. Research~\cite{li2021value} has been focused on probing different video representations such as ResNet~\cite{he2016deep}, C3D~\cite{hara2018can} and S3D~\cite{miech2020end} or CLIP-ViT~\cite{sun2023eva}, for improving video captioning quality. 

Dosovitskiy~\cite{dosovitskiy2020image} showed that the Transformer architecture, which has been initially developed for machine translation, can also be applied in computer vision tasks, outperforming CNNs in image classification tasks. From then on, Transformers have been successfully applied in a broad range of Computer Vision tasks including tasks performed on videos: action recognition~\cite{liu2022video}, video captioning~\cite{lin2022swinbert} or even multi-modal (vision and language) learning~\cite{fu2023empirical,chen2023valor,chen2023cosa}. VALOR~\cite{chen2023valor} uses three separate encoders for video, audio and text modalities and a single decoder for multi-modal conditional text generation. This architecture is pretrained on 1M audible videos with human annotated audiovisual captions, using multi-modal alignment and multi-modal captioning tasks. PDVC~\cite{wang2021end} frame the dense caption generation as a set prediction tasks with competitive results, compared to previous approaches based on the two-stage “localize-then-describe” framework.

Unified vision and language foundational models are either trained using both images and videos simultaneously~\cite{alayrac2022flamingo} or use a two-stage approach~\cite{wang2021simvlm, yu2022coca} in which the first stage contains image-text pairs, followed by a second stage in which video-text pairs are added. This two-stage approach has the advantage of faster training, models can be scaled up easier, and data is more freely available. VAST~\cite{chen2023vast} is a unified foundational model across three modalities: video, audio and text. To alleviate the limited scale and quality of video-text training data, COSA~\cite{chen2023cosa} converts existing image-text data into long-form video data. Then an architecture based on ViT~\cite{dosovitskiy2020image} and BERT~\cite{devlin2019bertpretrainingdeepbidirectional} is trained on this new long-form data. GIT~\cite{wang2022git} is a unified vision-language model with a very simple architecture consisting of a single image encoder and a text decoder, trained with the standard language modeling task. mPLUG-2~\cite{Xu2023mPLUG2AM} builds multi-modal foundational models using separate modules including video encoder, text encoder, image encoder followed by universal layers, a multi-modal fusion module and finally a decoder module.

What all these methods lack is the explainability factor, as the inner representation is opaque. Methods to obtain some of explainability include adding Reasoning Module Networks (RMNs) to guide the text generation process (e.g. for video captioning), including Explainable modules based on objects detected in saliency maps~\cite{szymanowicz2022discrete} or applying model agnostic techniques such as LIME~\cite{mishra2017local}.

Graph of Events in Space and Time - GEST~\cite{masala2023explaining} provides an explicit spatio-temporal representation of stories as they naturally appear in any median (e.g., videos, texts). GEST was previously shown to be a meaningful representation, providing a unified (vision and text) and explainable space in which semantic similarities can be effectively computed~\cite{masala2023explaining}. The main elements of GEST are events, represented as nodes and their interactions, in the form of edges. The nodes represent events, ranging from simple to more complex actions, constrained to a specific time and space. GEST edges relate two events and can define any kind of interaction, from temporal to semantic to logical. While previously used for generating videos in this work we implement the GEST concept the other way, starting from real videos towards GEST and finally a rich textual description. 

%% file: sec/3_method.tex
\section{Method}


To properly describe all kinds of videos ranging from simple to more complex (e.g., longer, with more actions and actors) you first have to analyze and understand what happens in a video. Furthermore, to ensure this process is explainable we decide to stray away from the current paradigm of sampling frames, processing, and feeding them into a model that builds an inner obfuscated numerical representation. Instead, we aim to harness the power and expressivity of Graphs of Events in Space and Time (GESTs)~\cite{masala2023explaining}. Therefore, our first goal is to understand the video, to build a pipeline that given an input video it automatically builds an associated GEST. Then, by reasoning over GEST we build an intermediate textual description in the form of a proto-language that is then converted to natural language description.

\begin{figure*}
\includegraphics[width=1.0\linewidth]{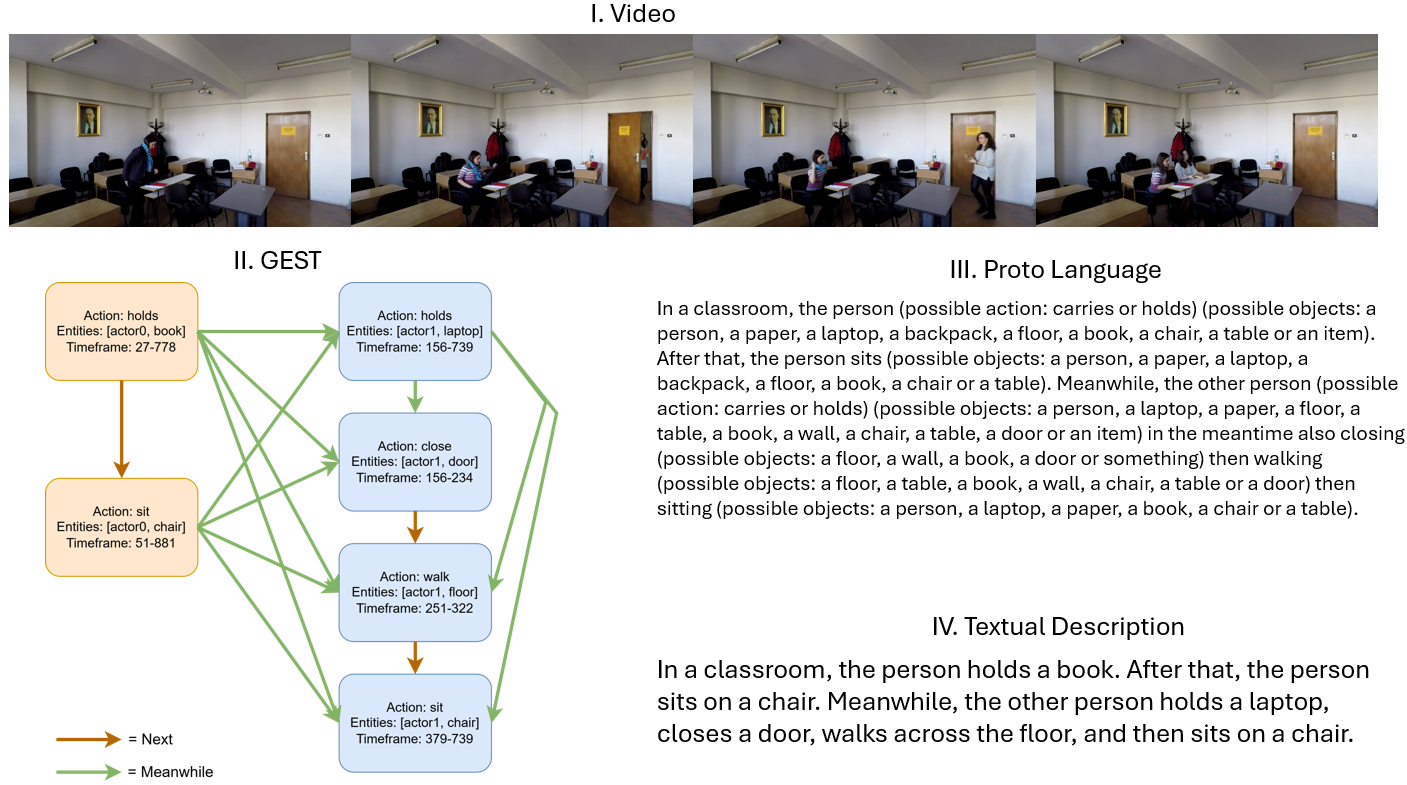}
\caption{A complete example of our proposed pipeline. Starting from the video, we automatically build the associated GEST. From this graph, we build the proto-language that is then fed to an LLM that generates the final textual description.}
\label{fig:complete_example}
\end{figure*}

In summary, our framework consists of two main steps: I. building the Graph of Events in Space and Time by processing and understanding frame level information, followed by reasoning to get an integrated, global view and II. translating this understanding in a rich natural language description by reasoning over GEST via a two-step process. For a complete example, starting from a video, building the GEST followed by the two-step process that generated the final description, see Figure~\ref{fig:complete_example}

\subsection{Understanding the video - Building the GEST}

In order to build an explicit representation of a video, we exploit existing sources of high-level image and video information: action detection, object detection and tracking, semantic segmentation and depth information. For each frame in a given video, we first extract this information followed by a matching and aggregation step. The output of the action detector includes, for every action a bounding box of the person performing the action together with the name of the action and a confidence score. Starting from this bounding box, we aim to gather all the objects in the vicinity of the person, objects that the actor could interact. First, the original bounding box is slightly enlarged to better capture the surroundings of the person, followed by finding all the objects that touch or intersect the new bounding box, based on information from the object detector and semantic segmentation. The list of objects is further filtered based on the intersection over union of the object and person bounding box with a fixed threshold, followed by depth-based filtering: we compute an average pixel-level depth for the person and the object, and if the depth difference between the person and the objects is between a set threshold, we consider the object close enough (both in "2D" based on intersection of bounding boxes and in "3D" based on depth) and we keep it in the list. All the objects that are not in the proximity of the person are discarded. Using this process, at this step we save for each action at each frame, information that includes the frame number, the person id (given at this point by the tracking model), the action name and confidence score as given the action detector, possibly involved objects and the bounding box of the person.

The next step is aggregating and processing frame-level information into global, video-level data. The first thing we noticed was that the model used for tracking had slight inconsistencies (e.g., changing the assigned id for a person from one frame to another even though the person in question did not move) or certain blind spots (e.g., losing sight of a person for a couple of frames). We noticed that a lot of the time the tracker would lose sight of a person for 5 to 10 consecutive frames. Upon detecting the person again it assigns a new id, as if it was a different person. We solve these short-term inconsistencies by unifying two person ids if they appear close in time (less than 10 frames) and they overlap enough (higher than 0.4 intersection over union). Note that these thresholds were set empirically by manually verifying around 20-25 examples. 

The lack of consistency of the tracker manifests itself both in short-term and long-term inconsistencies. An example of long-form inconsistency is when a person exists the frame, either due to camera movement or the person moving, and then re-enters the frame at a later time and in a different position. The previous solution can not work for this long-term inconsistency. Instead we are looking for semantic-based solution that is powerful enough for person re-identification while being very fast. For each person detected, in each frame we compute a feature vector based on the HSV histogram. For each pixel in the segmentation/mask of a person we bin the hue, saturation and value and linearize the resulting 3 dimensional space into a vector. We further compare such representations using cosine similarity. Finally, when a person appears in a frame, we compute its representation and compare it to previously seen persons. If the highest similarity exceeds a set threshold, we unify them into a single entity. 

The next step after person unification, is frame-based action filtering: based on empirically set thresholds, we filter our actions with confidence lower than 0.75 and for each frame keep only the two most confident actions. Then, to ensure a certain robustness, we implement a voting mechanism as follows: for each action in a frame we consider the previous five and the next five frames and if an action appears less than five times in this window of 11 frames, we discard it. This voting mechanism alleviates some of the inconsistencies and ensures a smoother action space. 

\begin{figure*}
\includegraphics[width=1.0\linewidth]{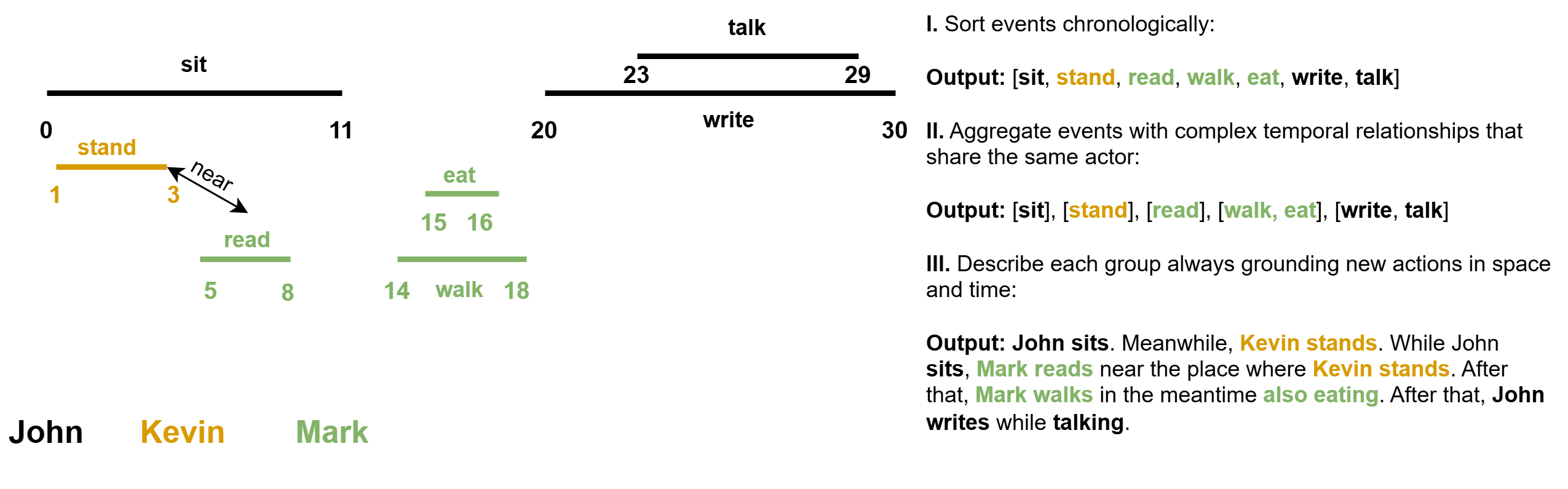}
\caption{On the left, an example of extracted events in space and time, with start and end frame. On the right, a high-level representation of the algorithm used for building the proto language.}
\label{fig:order}
\end{figure*}

Armed with this rich frame-level information we proceed to build the video-level representation. The first step is aggregating actions that appear in consecutive frames in events by saving the start and end frame ids, possible objects involved (union over objects at each frame, keeping objects that appear at least in 10\% of frames between start and end frame) and bounding boxes. Finally, we perform an additional unification step in which we aim to detect cases in which we find events with the same actors and the same action that are close in time (e.g. one starts at frame 10 and ends at frame 120 while the second starts at frame 130 and ends at frame 250) but are considered two different events. As such, we unify such events, again to make the final event-space less fragmented and more coherent. 

At this moment in time we have a list of events and for each event have actors, objects, timeframe (start and end frame ids) and location (bounding boxes). The last step in this entire pipeline in building spatio-temporal relationships between events. As both temporal and spatial information is readily available for each event, this is a rather straightforward process: we build pairs of events and if they meet certain criteria we link them in space or time. For spatial relations between two events that have an overlap in time, for each such frame, we are interested in the two actions being close in space. Therefore we compute the ratio between the Euclidean distance of the centroids and the sum of the diagonals and if this ratio is lower than a certain threshold we consider that the two actions are related (i.e., close) in space. If this happens for more than 75\% of the overlapping frames we consider the events to be close in space and mark them accordingly (i.e. build an edge, a spatial relation in the graph between the two events). For temporal relations we follow a similar approach, we are checking pair of events, characterize three types of temporal relations: next, same time and meanwhile. 

This leaves us with an over-complete graph, as it contains an over-complete set of possible objects for each event. Better grounding and obtaining a concrete GEST can be obtained in a variety of ways including picking objects based on proximity to the person or by the "temporal" size (number of frames in which is close to the person). We solve this at a later stage in the pipeline, by allowing an LLM to pick the most probable object. For more details see the following section.

For action detection, we use VideoMAE~\cite{tong2022videomae} finetuned on AVA-Kinetics~\cite{li2020ava}. Object detection and tracking are performed using the YOLO pipeline~\cite{yolov8_ultralytics}, while semantic segmentation is performed using Mask2Former~\cite{cheng2021mask2former}. Finally, Marigold~\cite{ke2023repurposing} is used to compute depth estimation.

\subsection{Generating a natural language description}

Translating a GEST into a cohere, rich and natural language description is not a straight-forward task with multiple possibilities. In this work we adopt a two-stage approach that harnesses the power of existing text-based LLMs to build natural descriptions. The goal of the first step in our approach in to convert the graph into sound but maybe a rough around the edges textual form, an initial description that we call proto-Language. While this representation is sound and accurately depicts the information encoded in the graph, it lacks a certain naturalness, as it may sound too robotic, lacking a more nuanced touch. Therefore, to obtain a more human-like description we use existing LLMs by feeding them with this proto-Language and prompting with the goal of rewriting the text to make it sound more natural.

The visual information is already converted and integrated into the GEST, but the question of how this graph can be effectively converted to an input to be consumed by an LLM still remains. The first step in this process involves a temporal sorting of the graph (by the start frame of each event; akin to a topological sort). If at each moment in time a single actor performs a single action, this is a rather straightforward process, with the results being a tree in space and time. With multiple actors and/or actions, this becomes more complex, with more than one possible representation. Our approach aggregates chronologically sorted actions into higher-level groups of actions by actors. Each such group is then described in text, by describing each event using a simple grammar and taking into account the intra-group and inter-group spatial and temporal relations. A high-level example of this algorithm is presented in Figure~\ref{fig:order}. Describing a single event involves describing the actor or actors (including objects) involved, the action performed, and spatial and temporal information if available.

\begin{figure}
\includegraphics[width=1.0\linewidth]{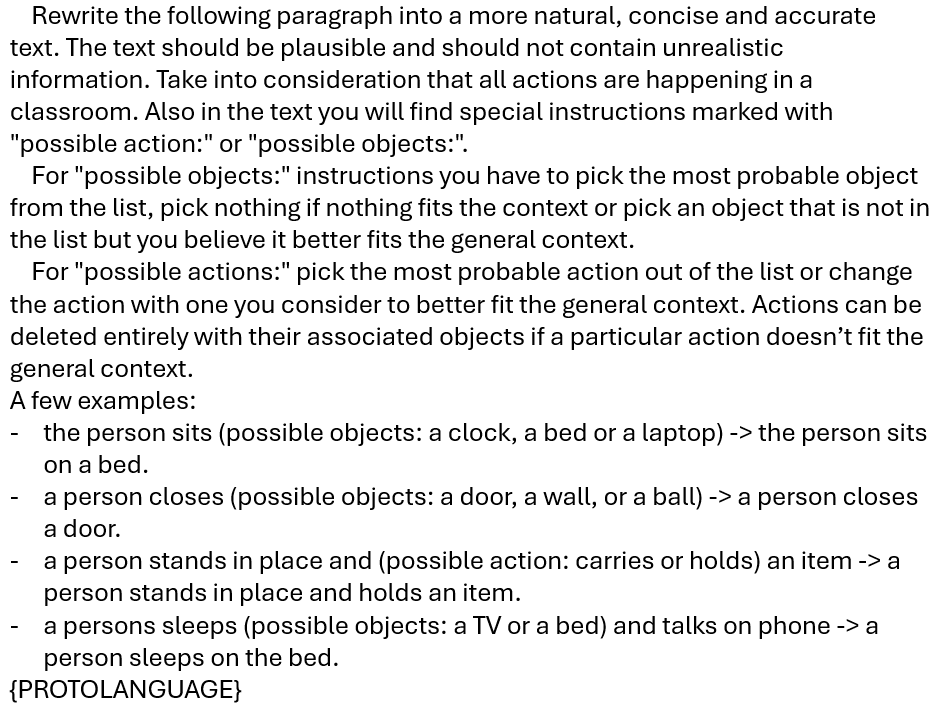}
\caption{The prompt used for generating the final text description.}
\label{fig:prompt}
\end{figure}

Crucially, we decide to not make a hard decision when selecting the possible objects involved in an event and to double down on the power of LLMs, feeding them with special instructions for selecting the most probable object in the given context. Therefore, when describing an event, we list all possible objects (as computed earlier)  and let the LLM pick the objects that are most probable to appear in the given context, with the power to pick a new object that is not present in the list or not pick an object at all. Furthermore, we allow the LLM to change the name of an action or delete an action and its associated entities entirely if it does not fit the context. The prompt and instructions used to generate the final text description are depicted in Figure~\ref{fig:prompt}.

Finally, to get a better understanding of the context we prompt a small vision language model\footnote{https://huggingface.co/vikhyatk/moondream2 last accesed on 9th of January 2025} with the following instruction: "In what scene does the action take place? Simply name the scene with no further explanations. Use very few words, just like a classification task, e.g., classroom, park, football field, mountain trail, living room, street." and prepend the answer to the proto language. This allows the LLM to better understand the context of the actions and objects and thus better ground the description in the real world.

%% file: sec/4_experiments.tex
\section{Experiments and Evaluation}

In this section, we describe the experimental settings, ranging from datasets to methods used and the selected evaluation methodology.


\subsection{Datasets}
To validate our approach, we employ five different datasets: Videos-to-Paragraphs~\cite{bogolin2020hierarchical}, COIN~\cite{tang2019coin}, WebVid~\cite{bain2021frozen}, VidOR~\cite{shang2019annotating} and ImageNet-VidVRD~\cite{shang2017video}.

\begin{figure}
\includegraphics[width=1.0\linewidth]{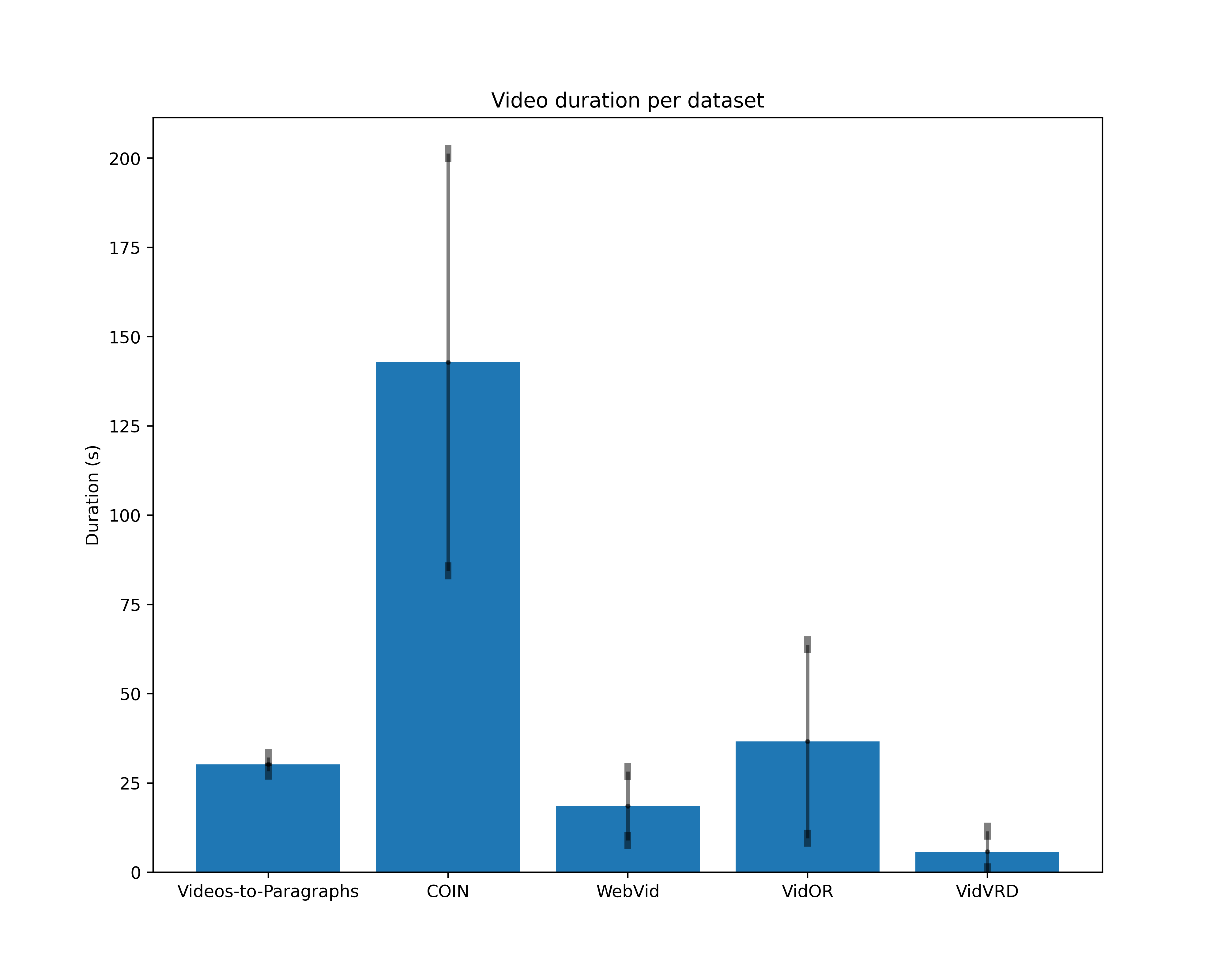}
\caption{Video duration statistics per dataset.}
\label{fig:video_length}
\end{figure}

Videos-to-Paragraphs~\cite{bogolin2020hierarchical} consists of 510 videos of actions performed by actors in a school-like environment, filmed with both moving and fixed cameras. All the videos contain a multitude of actions including interactions between two or more actors. The complexity of the Videos-to-Paragraphs videos stems rather from the multitude of actors and actions rather than from the complexity of individual actions. The COIN dataset~\cite{tang2019coin} consists of over 11k videos of people solving 180 different everyday tasks in 12 domains (e.g., automotive, home repairs). All videos were collected from Youtube\footnote{www.youtube.com last accessed on 30th December 2024}, with an average duration of 2.36 minutes. We chose this data set for its rather long and complex nature. VidVRD~\cite{shang2017video} and VidOR~\cite{shang2019annotating} consist of 1k and 10k video annotated with visual relations. VidVRD contains 35 unique subject/object categories with a total of 132 predicate categories. Similarly, in VidOR 80 categories of objects and 50 categories of relations are annotated. We select video from both sources for their rich visual relations, often containing multiple actors performing a multitude of complex intertwined actions. WebVid~\cite{bain2021frozen} contains 10 million rich and diverse web-scraped videos with short text descriptions. We pick videos from this dataset mainly for the diverse base that it offers (e.g. a wide range of possible actions and environments). For each dataset, video duration statistics are presented in Figure~\ref{fig:video_length}.

At this point, it is important to make a clear distinction between the types of videos in the five datasets and why we consider the Videos-to-Paragraphs dataset the most relevant for the task at hand (i.e., rich video description). The reason is two fold: on one side Videos-to-Paragraphs dataset contains rich, two-level human annotated descriptions (i.e., SVO and video level descriptions) so it represents a strong benchmark. Secondly, the dataset was built in such a way that each video has a clear context, there are a lot of interactions with objects and between persons, and crucially, there is no single encompassing action that could properly describe the actions performed in the video. While Videos-to-Paragraphs videos are not the longest, they are in this sense the most complex. Instructional videos, such as COIN videos, that are significantly longer by definition could be described with great accuracy by their overarching action (e.g., teaching how to install parquet). Similarly for the other datasets, they are built for and defined by the action happening in said video. Furthermore, we did not find any evidence that any of the considered methods has been trained on this dataset which is not the case for other considered dataset (e.g., VALOR has been trained on a combination of datasets that also contains WebVid). This makes Videos-to-Paragraphs dataset an even stronger choice, as we can almost guarantee that is has not been used at training time for any method and thus can be considered a novel, even out-of-distribution dataset.


\subsection{Methods}
We compare our approach (GEST) against a suite of existing open models: VidIL~\cite{wang2022language}, VALOR~\cite{chen2023valor}, COSA~\cite{chen2023cosa}, VAST~\cite{chen2023vast}, GIT2~\cite{wang2022git}, mPLUG-2~\cite{Xu2023mPLUG2AM} and PDVC~\cite{wang2021end}. Upon careful inspection of generated texts, we found that VidIL generated texts tend to be rich, but contain a high degree of hallucinations, while descriptions generated by our method tend to miss certain relevant aspects; see Section~\ref{sec:observations} for more details. Grounding VidIL and vice versa, adding more details to our approach should increase the overall quality of the descriptions. Therefore we add the output of our method to the input used by VidIL (e.g. frame captions, events) and re-run GPT 4o to generate a textual description. Thus, the only changes we apply to VidIL are simply adding the textual description generated by our approach to the set of inputs already used by VidIL and minimally tweaking the generation prompt.

\subsection{Evaluation}

To evaluate our approach and compare it with existing models, we use two evaluation protocols. On one hand, we turn to a text-based evaluation based on standard text similarity metrics (akin to how captioning methods are evaluated), while on the other hand, we perform a study to obtain qualitative ranking of the generated texts. 

While for videos in Videos-to-Paragraphs~\cite{bogolin2020hierarchical} we have access to a rich, narrative-like ground truth, for the other datasets this is not readily available. Therefore, we use GPT 4o to generate pseudo-ground truth rich descriptions. For the ranking part, we harness strong Vision Large Language Models (VLLMs) and faced with a video and six automatically generated texts, their goal is to rank the videos from best to worst based on richness and factual correctness. We selected the methods used for this evaluation using the quantitative results already obtained and by running a very small-scale initial experiment with a human annotator. In the end, the survey includes texts generated by the following methods: GEST (own), VidIL, GIT2, mPLUG-2, PDVC and GEST (own) + VidIL. We implement the LLM-as-a-Jury~\cite{verga2024replacing} approach, with Claude 3.5\footnote{claude-3-5-sonnet-20241022}, GPT 4o\footnote{gpt-4o-2024-11-20}~\cite{hurst2024gpt}, and Qwen2\footnote{Qwen2-VL-72B-Instruct}~\cite{Qwen2VL} prompted with 10 uniformly sampled frames from each video, the six generated descriptions and the set of instructions. Beyond the ranking, we prompt the VLLMs to also provide a score between 1 and 10, to better understand the differences between the methods.

%% file: sec/5_results.tex
\section{Results}

Finally, we present some qualitative examples and highlight some patterns observed throughout a multitude of videos and generated descriptions.

\begin{figure*}
\includegraphics[width=1.0\linewidth]{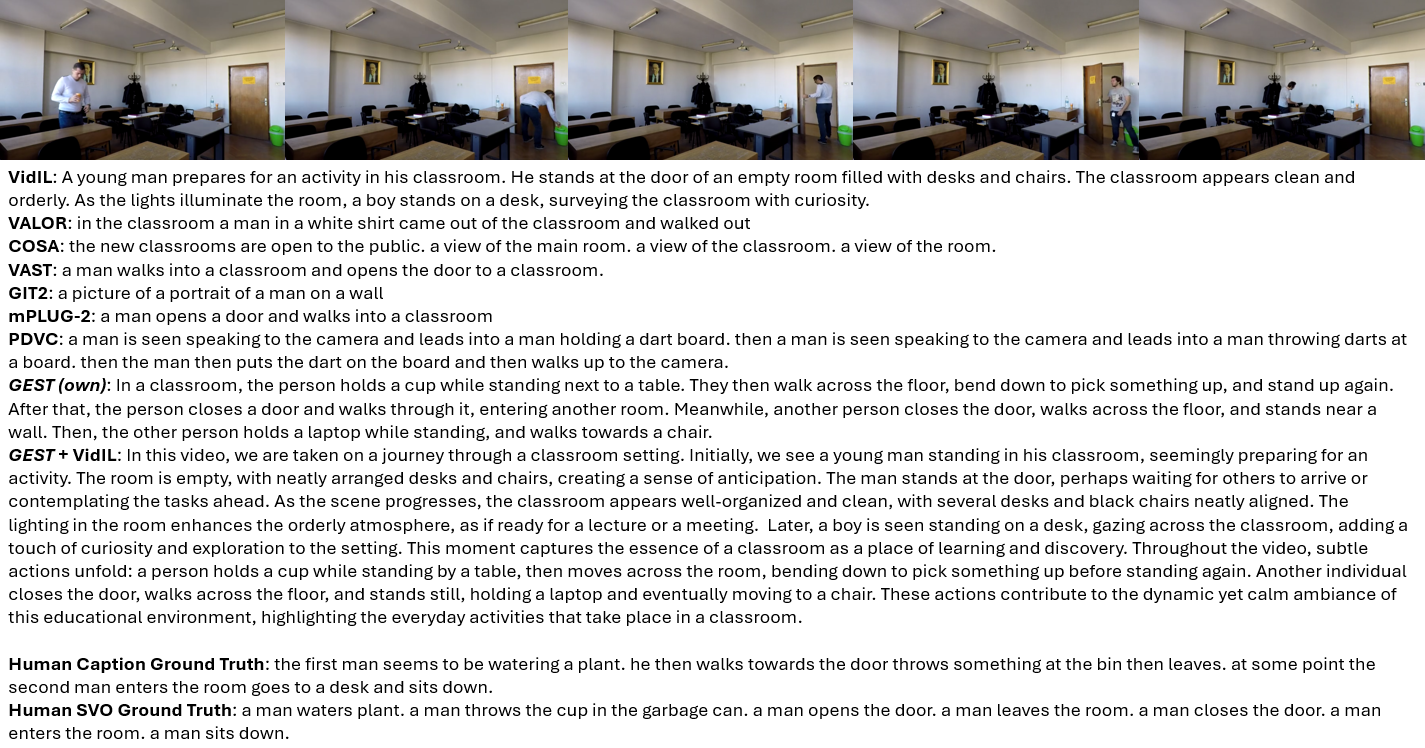}
\caption{Video descriptions generated by all considered methods together with ground truth for a video in Videos-to-Paragraphs dataset. Note that other methods completely miss the two different persons in the video, a crucial element. Most of the methods describe only the second person entering the room, completely missing the first person's actions. Our method correctly identifies that there are two distinct persons in the video and describes most of the actions in the videos, missing the first action due to the plant being out of frame and the last action due to video ending prematurely.}
\label{fig:full_example_captions}
\end{figure*}

\subsection{Quantitative Evaluation - Captioning metrics}

For the Videos-to-Paragraphs dataset, where rich ground truth is available, we present results for both levels of annotations available (i.e., caption and SVO-level) in Table~\ref{tab:imar-caption-longest} and Table~\ref{tab:imar-svos-longest}. Note that in both cases, our proposed method performs the best. For this dataset the combination of our method with VidIL is underperforming, when compared with captions it is performing worse on average than the two methods individually, while for the SVO-level description it slightly improves over VidIL but trails behind our method. 

The results for the five considered datasets are aggregated in Table~\ref{tab:metrics-full} while per dataset averages are presented in Table~\ref{tab:metrics-per-categ}. For datasets besides Videos-to-Paragraphs, VidIL performs significantly better than other considered methods. This is in part due to the nature of the videos and the nature of the GPT generated pseudo-ground truth that contain a lot of details, details that are captured by the input extraction methods used by VidIL. Other methods tend to focus more on describing the action and less on describing details of the scene (e.g., how people are dressed). Combining these rich details about the scene with a rich description of the actions performed in the video leads to more qualitative descriptions, as the results prove: the combination of our method with VidIL obtains top scores on all datasets with the exception of VidVRD where it obtains a competitive results, very close to the top performer. 


\begin{table*}
  \centering
  \begin{tabular}{lc|ccccccc}
    \toprule
    Method & Average  & Bleu@4 & METEOR & ROUGE-L & CIDEr & SPICE & BERTScore & BLEURT \\
    \midrule
    VidIL~\cite{wang2022language} & 13.24 & 0.76 & 9.95	& 18.72	& 1.69	& 10.08 &	10.99& 	40.50 \\
    VALOR~\cite{chen2023valor} & 12.38 & 0.35 &	5.24&	16.02&	1.41	&11.89	&16.59	&35.16\\
    COSA~\cite{chen2023cosa} &11.45 & 1.16&	6.58&	20.19&	\textbf{3.56}&	7.76&	0.15&	40.75\\
    VAST~\cite{chen2023vast} & 14.88 & 0.58&	6.11&	18.89	&1.59&	13.73&	\textbf{22.44}&	40.83\\
    GIT2~\cite{wang2022git} & 13.61 & 0.23&	4.83&	16.34&	1.54&	11.99&	20.35&	39.96\\
    mPLUG-2~\cite{Xu2023mPLUG2AM} & 12.14 & 0.03&	3.65&	11.89&	0.42&	\textbf{14.07}	&18.32	&36.59\\
    PDVC~\cite{wang2021end}& 14.18 & 1.14 &	10.92 &	\textbf{24.02} &	1.85 &	9.98 &	7.25 &	44.10\\
    \textit{GEST (own)} & \textbf{15.05} & \textbf{1.19}&	12.06&	20.76&	2.84&	8.32&	15.71	&44.47\\
    \midrule
    \textit{GEST} + VidIL~\cite{wang2022language} & 12.12 & 0.57&	\textbf{12.80}&	15.14&	0.00&	7.34&	3.00&	\textbf{45.97} \\
    \bottomrule
  \end{tabular}
  \caption{Videos-to-Paragraphs results when using the longest human annotated caption as ground truth. \textbf{Bold} marks the best result in each category. Note that our method GEST is the top performer with competitive results on most of the considered metrics, being first or second on five of the seven metrics.}
  \label{tab:imar-caption-longest}
\end{table*}

\begin{table*}
  \centering
  \begin{tabular}{lc|ccccccc}
    \toprule
    Method & Average  & Bleu@4 & METEOR & ROUGE-L & CIDEr & SPICE & BERTScore & BLEURT \\
    \midrule
    VidIL~\cite{wang2022language} & 11.16 & 1.11&	7.38&	18.88	&1.70	&9.27&	-2.40	&42.18\\
    VALOR~\cite{chen2023valor} & 9.37 & 0.06&	3.34&	12.87&	0.92&	10.42&	-0.56	&38.56\\
    COSA~\cite{chen2023cosa} & 12.05 & 0.81&	5.25&	21.38&	1.10	&5.93	&4.50&	45.34\\
    VAST~\cite{chen2023vast} & 11.34 & 0.15&	3.77&	14.03	&1.03&	\textbf{12.94}&	3.71&	43.76\\
    GIT2~\cite{wang2022git} & 10.43 & 0.03&	2.85&	11.32&	0.93&	10.84&	2.18&	44.87\\
    mPLUG-2~\cite{Xu2023mPLUG2AM} & 9.86 & 0.00&	2.24&	8.82&	0.24&	12.46&	1.50&	43.76\\
    PDVC~\cite{wang2021end} & 13.34 & 0.45&	7.79&	\textbf{22.99}&	1.54&	9.06&	6.01&	45.53\\
    \textit{GEST} & \textbf{13.59}& \textbf{1.41}&	10.03&	20.50&	\textbf{1.75}&	8.54&	\textbf{6.91}&	45.98\\
    \midrule
    \textit{GEST} + VidIL~\cite{wang2022language} &11.81 & 0.73	&\textbf{12.49}&	18.89&	0.01&	6.81&	-5.17&	\textbf{48.93}\\
    \bottomrule
  \end{tabular}
  \caption{Videos-to-Paragraphs results when using the longest SVO-level human annotation as ground truth. \textbf{Bold} marks the best result in each category. As in Table~\ref{tab:imar-caption-longest}, we note that our method GEST is the top performer on average, with competitive performance on most of the metrics.}
  \label{tab:imar-svos-longest}
\end{table*}

\begin{table*}
  \centering
  \begin{tabular}{lc|ccccccc}
    \toprule
    Method & Average  & Bleu@4 & METEOR & ROUGE-L & CIDEr & SPICE & BERTScore & BLEURT \\
    \midrule
    VidIL~\cite{wang2022language} &16.04 & 2.65&	10.41&	20.87&	\textbf{6.28}	&\textbf{12.63}	&\textbf{16.15}&	43.29	\\
    VALOR~\cite{chen2023valor} &7.95 &0.02&	3.04&	9.86&	0.18&	7.52&	6.78&	28.25\\
    COSA~\cite{chen2023cosa} & 10.37 & 0.82&	6.32&	17.33	&0.29&	8.76&	2.45&	36.65\\
    VAST~\cite{chen2023vast} & 10.54 & 0.07&	3.84&	12.77&	0.21	&9.80&	13.20&	33.88\\
    GIT2~\cite{wang2022git} & 10.55 & 0.03&	3.57&	11.91&	0.19&	9.83&	12.57&	35.73\\
    mPLUG-2~\cite{Xu2023mPLUG2AM} & 8.38 & 0.00&	2.42&	8.01&	0.05&	8.92&	8.23&	31.03\\
    PDVC~\cite{wang2021end} & 11.70 & 0.74&	7.49&	21.07	&0.43	&6.86&	4.07	&41.25\\
    \textit{GEST} & 11.63 & 0.99 &	7.52&	17.83&	0.84&	7.04&	7.18&	40.00\\
    \midrule
    \textit{GEST} + VidIL~\cite{wang2022language} & \textbf{17.54} & \textbf{3.07}&	\textbf{17.65}	&\textbf{23.09}	&0.13&	12.19	&15.69&	\textbf{50.93}\\
    \bottomrule
  \end{tabular}
  \caption{Aggregated (per dataset; Videos-to-Paragraphs, COIN, WebVid, VidOR, VidVRD) quantitative results. \textbf{Bold} marks the best result in each category. Note the very strong performance of the combination of GEST and VidIL, together with VidIL. The two methods perform significantly better than the other methods on average and on most metrics. }
  \label{tab:metrics-full}
\end{table*}

\begin{table*}
  \centering
  \begin{tabular}{lc|ccccc}
    \toprule
    \multirow{2}{*}{Method} & Average & VtP & COIN & WebVid & VidOR & VidVRD\\
    & & (489) & (75) & (92) & (93) & (62)\\
    \midrule
    VidIL~\cite{wang2022language} & 16.04 & 11.16 & 14.66 & 18.19 & 16.93 & \textbf{19.36}\\
    VALOR~\cite{chen2023valor} & 7.95 & 9.37 & 5.03 & 7.40 & 9.08 & 8.87\\
    COSA~\cite{chen2023cosa} & 10.37 & 12.05 & 7.56 & 9.44 & 10.59 & 12.24\\
    VAST~\cite{chen2023vast} & 10.54 & 11.34 & 7.16 & 12.93 & 10.66 & 10.60\\
    GIT2~\cite{wang2022git} & 10.55 & 10.43 & 9.22 & 12.33 & 10.69 & 10.06\\
    mPLUG-2~\cite{Xu2023mPLUG2AM} & 8.38 & 9.86 & 5.47 & 9.84 & 8.48 & 8.25\\
    PDVC~\cite{wang2021end} & 11.70 & 13.34 & 10.63 & 11.48 & 11.63 & 11.42\\
    \textit{GEST} & 11.63 & \textbf{13.59} & 9.96 & 11.40 & 12.51 & 10.69\\
    \midrule
    \textit{GEST} + VidIL~\cite{wang2022language} & \textbf{17.54} & 11.81 & \textbf{16.84} & \textbf{20.29} & \textbf{19.43} & 19.31\\
    \bottomrule
  \end{tabular}
  \caption{Average over text similarity metrics for each dataset. VtP - Videos-to-Paragraphs. \textbf{Bold} marks the best result. For each dataset we note the number of videos used. Again, we note the top performing method is the combination of our method, GEST, with VidIL. For three out of the five datasets, it obtains the highest score, while for Videos-to-Paragraphs and VidVRD the top performing methods are GEST and VidIL respectively. }
  \label{tab:metrics-per-categ}
\end{table*}


\subsection{Qualitative Evaluation - Method Ranking}

The results are presented in Tables~\ref{tab:eval-vllm-rank} and~\ref{tab:eval-vllm-grade}. Again, we note the very strong performance of our method on Videos-to-Paragraphs dataset, with the lowest rank and highest grade at a significant distance from other methods. Out of the considered methods, GIT2 and mPLUG-2 have by far generated the shortest and "simplest" descriptions (akin to video captioning) and their similarity is clearly seen in the results: they are very close both when considered the rank and the grades. This is in somewhat contrast to the quantitative results, where the differences between the two methods are clearly visible. PDVC, a competitive method if judged by text similarity metrics, is clearly underperforming if qualitatively judged. Combining our method with VidIL tends to increase the overall quality of the generated texts, obtaining a better ranking on 3 out of the 5 datasets, with small differences on the other 2.

\subsection{Qualitative Examples and Observations}
\label{sec:observations}

We present a sample video together with all the generated descriptions and ground truth in Figure~\ref{fig:full_example_captions}. By manually investigating more than 200 videos with their associated descriptions we noticed some strong patterns: both GIT2 and mPLUG-2 method generated very short descriptions in the form of one sentence, mentioning a single entity (that could include more than one actor e.g., "two men") and a single action. These descriptions are very simple, trivially true (a sentence that only describes the surroundings, or a sentence that states that a person is somewhere) and most of the time completely miss actions and actors. This makes them suitable for videos which have a single overarching action. 

While arguably competitive based on qualitative metrics, PDVC-generated descriptions are too scriptic and contain way too little information to be relevant in real-world scenario. This proves yet again that automatic evaluation based on text similarity metrics is not the be-all end-all solution for evaluating video descriptions as our analysis casts a serious doubt on the effectiveness of such an approach. 

On the other side, descriptions generated by VidIL are far richer, in some cases too rich, containing a lot of hallucinations and untrue facts. For example, in most Videos-to-Paragraph samples, as it sees a person in a room with a chalkboard it automatically infers that that person is a teacher, even if the person is sitting alone in the room, at a desk, doing something completely unrelated to teaching. It even hallucinates non-existing students (for some reason always six students) that are attentive to this imagined teacher, even if in the entire video there is only one person. Also, if a person is holding or writing on a laptop, that person "becomes" a computer scientist and all of the subsequent actions are described through this new persona (e.g. writing on laptop becomes coding). 

As our method is based on an action recognizer that has a rather small and fixed set of possible actions, our generated descriptions lack flexibility and sometimes exhibit a limited understanding of the world. They tend to describe lower-level actions, for example, mopping the flooring might be described by holding an object while walking around. This also explains the strong performance of our method on Videos-to-Paragraphs dataset as the videos complexity stems from the multitude actions and interactions between multiple actors, rather than from individual action complexity. 

Combining our method with VidIL yields mixed results: in some cases the generated description is more grounded, containing fewer hallucinations, while in other cases our input seems to be irrelevant. This seems to happen more often where the exists a strong disagreement between the two types of input (e.g., number of persons present in the video).


\begin{table*}
  \centering
  \begin{tabular}{lc|ccccc}
    \toprule
    \multirow{2}{*}{Method} & Average & VtP & COIN & WebVid & VidOR & VidVRD\\
    & & (489) & (75) & (92) & (93) & (62)\\
    \midrule
    VidIL~\cite{wang2022language} & 2.88 & 3.49 & 2.55 & \textbf{2.71} & 2.84 & \textbf{2.84}\\
    GIT2~\cite{wang2022git} & 3.44 & 3.47  & 3.53 & 3.24 & 3.60 & 3.33 \\
    mPLUG-2~\cite{Xu2023mPLUG2AM} & 3.44 & 3.60 & 3.21 & 3.46 & 3.64 & 3.32\\
    PDVC~\cite{wang2021end} & 5.27 & 5.60 & 5.06 & 5.46 & 5.03 & 5.21\\
    \textit{GEST} & 3.16 & \textbf{1.79} & 4.20 & 3.11 & 3.25 & 3.42  \\
    \midrule
    \textit{GEST} + VidIL & \textbf{2.81} & 3.05 & \textbf{2.45} & 3.02 & \textbf{2.64} & 2.89\\
    \bottomrule
  \end{tabular}
  \caption{Average rank (best is 1, worst is 6) as selected by VLLMs. VtP - Videos-to-Paragraphs. \textbf{Bold} marks the best result in each category. The top performing method as evaluated using the LLM-as-a-Jury approach is again the combination between our method GEST and VidIL.}
  \label{tab:eval-vllm-rank}
\end{table*}

\begin{table*}
  \centering
  \begin{tabular}{lc|ccccc}
    \toprule
    \multirow{2}{*}{Method} & Average & VtP & COIN & WebVid & VidOR & VidVRD\\
    & & (489) & (75) & (92) & (93) & (62)\\
    \midrule
    VidIL~\cite{wang2022language} & 5.98 & 4.99 & \textbf{6.54} & \textbf{6.20} & 6.08 & \textbf{6.12} \\
    GIT2~\cite{wang2022git} & 4.96 & 4.58 & 4.92 & 5.24 & 4.79 & 5.28 \\
    mPLUG-2~\cite{Xu2023mPLUG2AM} & 4.96 & 4.50 & 5.25 & 5.07 & 4.74 & 5.31 \\
    PDVC~\cite{wang2021end} & 2.85 & 2.23 & 3.14 & 2.63 & 3.22 & 3.04 \\
    \textit{GEST} & 5.60 & \textbf{7.30} & 4.30 & 5.60 & 5.50 & 5.32 \\
    \midrule
    \textit{GEST} + VidIL & \textbf{6.02} & 5.55 & 6.46 & 5.81 & \textbf{6.25} & 6.00\\
    \bottomrule
  \end{tabular}
  \caption{Average grade (best is 10, worst is 1) as selected by VLLMs. VtP - Videos-to-Paragraphs. \textbf{Bold} marks the best result in each category. \textbf{Bold} marks the best result in each category. The top performing method as evaluated using the LLM-as-a-Jury approach is again the combination between our method GEST and VidIL.}
  \label{tab:eval-vllm-grade}
\end{table*}


%% file: sec/6_conclusions.tex
\section{Conclusions}

We have proposed a novel method that combines state-of-the-art models from both computer vision and natural language processing domains with a procedural module to generate explainable video descriptions. It uses object and action detectors, semantic segmentation and depth estimation to automatically extract frame-level information, that is further aggregated into video-level events, ordered in space and time. Using a relatively simple algorithm, events and their spatio-temporal relations are further converted into a proto-language that is rich in information, but lacks fluency and grammatical complexity. Using LLMs, this simple language is finally converted into a fluent, coherent story that describes the events in natural language. To our best knowledge, we are the first to explore such a procedural approach, that bridges existing state of the art learning models from vision and language in order to provide an explainable solution to the long-standing vision to language translation problem. Our experiments on videos from several current datasets, show that our zero-shot approach can outperform the current state of the art open models that are heavily trained for video captioning.

Furthermore, our method greatly outperforms existing methods on Videos-to-Paragraphs dataset, a grounded and complex dataset with multiple actions and actors. Our approach is especially suited for this kind of videos, for example surveillance and security videos, accurately describing the actors and actions performed.